\documentclass[conference]{IEEEtran}
\IEEEoverridecommandlockouts

\usepackage{cite}
\usepackage[numbers,sort&compress]{natbib}
\usepackage{amsmath,amssymb,amsfonts}
\usepackage{algorithmic}
\usepackage{graphicx}
\usepackage{textcomp}
\usepackage{xcolor}
\usepackage{multirow}
\usepackage[misc]{ifsym}
\usepackage{booktabs}

\usepackage{makecell}
\usepackage{tabularx}
\usepackage{subfigure}
\usepackage{color}
\usepackage{xspace}
\usepackage{enumitem}
\usepackage{ragged2e}
\usepackage{bbding}
\usepackage{chngcntr} 

\definecolor{citecolor}{HTML}{1F801F}
\definecolor{linkcolor}{HTML}{ED1C24}
\usepackage[colorlinks,citecolor=citecolor, linkcolor=linkcolor, bookmarks=false]{hyperref}

\newcommand{\loose}[0]{\looseness=-1}

\newlength\savewidth\newcommand\shline{\noalign{\global\savewidth\arrayrulewidth
  \global\arrayrulewidth 1.5pt}\hline\noalign{\global\arrayrulewidth\savewidth}}

\def\BibTeX{{\rm B\kern-.05em{\sc i\kern-.025em b}\kern-.08em
    T\kern-.1667em\lower.7ex\hbox{E}\kern-.125emX}}
\begin{document}

\def\x{{\mathbf x}}
\def\L{{\cal L}}

\let\OLDthebibliography\thebibliography
\renewcommand\thebibliography[1]{
  \OLDthebibliography{#1}
  \setlength{\parskip}{0pt}
  \setlength{\itemsep}{0pt plus 0.3ex}
}

\newcommand{\methodname}{SIAM\xspace}
\newcommand{\blockname}{DaMi\xspace}
\renewcommand{\paragraph}[1]{\noindent\textbf{#1}\quad}

\title{\methodname: A Simple Alternating Mixer \\ for Video Prediction\\
\thanks{\hrule
\vspace{0.5ex}
\textsuperscript{\dag}Corresponding author.}}

\author{
    \IEEEauthorblockN{Xin Zheng$^1$ \qquad Ziang Peng$^2$ \qquad Yuan Cao$^1$ \qquad Hongming Shan$^3$ \qquad Junping Zhang$^{1, \dag}$
    \thanks{This work is supported by National Natural Science Foundation of China~(NSFC 62176059). The computations in this research were performed using the CFFF platform of Fudan University.}}
    \IEEEauthorblockA{$^1$ Shanghai Key Lab of Intelligent Information Processing, School of Computer Science}
    \IEEEauthorblockA{$^2$ Department of Aeronautics and Astronautics}
    \IEEEauthorblockA{$^3$ Institute of Science and Technology for Brain-inspired Intelligence}
    \IEEEauthorblockA{Fudan University, Shanghai 200433, China}
    \IEEEauthorblockA{\{xzheng22, zapeng21\}@m.fudan.edu.cn, \{caoy16, hmshan, jpzhang\}@fudan.edu.cn}
}

\maketitle

\begin{abstract}
Video prediction, predicting future frames from the previous ones, has broad applications such as autonomous driving and weather forecasting. Existing state-of-the-art methods typically focus on extracting either spatial, temporal, or spatiotemporal features from videos. Different feature focuses, resulting from different network architectures, may make the resultant models excel at some video prediction tasks but perform poorly on others. Towards a more generic video prediction solution, we explicitly model these features 
in a unified encoder-decoder framework and propose a simple alternating Mixer (\methodname). The novelty of \methodname lies in the design of dimension alternating mixing (\blockname) blocks, which can model spatial, temporal, and spatiotemporal features through alternating the dimensions 
of the feature maps. Extensive experimental results demonstrate the superior performance of the proposed \methodname on four benchmark video datasets covering both synthetic and real-world scenarios.
\end{abstract}


\section{Introduction}
\label{sec:intro}

The ability to predict future frames by learning from historical contents provides visual clues for humans 
and artificial intelligent systems to make judicious decisions in various situations, such as 
autonomous driving~\cite{voletiMCVDMaskedConditional2022}, weather forecasting~\cite{caoMutualInformationBased2023a}, 
and robotic control~\cite{finnUnsupervisedLearningPhysical2016a}. 
Due to its wide range of applications, video prediction has attracted growing research interests 
recently~\cite{opreaReviewDeepLearning2022}. However, it remains challenging to develop generic video prediction models 
with the capability to learn complex dynamics, \textit{e.g.}, motions and deformations, 
while maintaining pixel-level details for high-resolution synthesis at the same 
time~\cite{guenDisentanglingPhysicalDynamics2020,wuMotionRNNFlexibleModel2021,changSTRPMSpatiotemporalResidual2022}.

\begin{figure}[h]
    \centering
    \subfigure[RNN-based models.]{\label{fig:cate-a} \includegraphics[width=\linewidth]{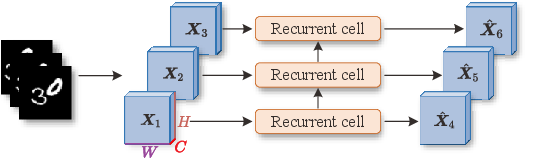}}
    
    \vfill
    \subfigure[2D-CNN-based models.]{\label{fig:cate-b} \includegraphics[width=\linewidth]{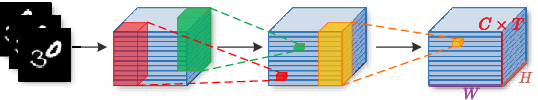}}
    
    \subfigure[ViViT- and 3D-CNN-based models.]{\label{fig:cate-c} \includegraphics[width=\linewidth]{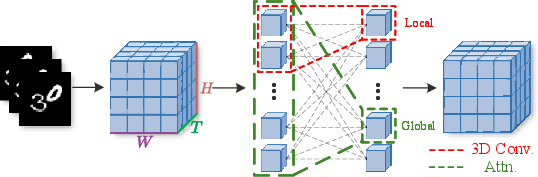}}

    \caption{Comparison of different architectures for handling video data. (a)~A sequence of frames are passed through RNNs, 
    and one prediction is made at each timestep. (b)~2D-CNNs manage videos the same as images by stacking all input 
    frames over the channel dimension. (c)~ViViT- and 3D-CNN-based models operate on patches in space and time simultaneously: ViViTs leverage the attention mechanism to process all patches globally, while 3D CNNs have relatively local receptive fields.}

    \label{fig:cate}
  \end{figure}

  Towards this end, prior works have developed novel operations and architectures. 
  As illustrated in Fig.~\ref{fig:cate-a}, the most commonly-used backbones are the recurrent 
  neural networks~(RNNs), where consecutive frames are passed through a recurrent 
  model~\cite{shiConvolutionalLSTMNetwork2015b, guenDisentanglingPhysicalDynamics2020, wangPredRNNRecurrentNeural2023}. 
  Recent RNN-based models typically prioritize innovative designs of recurrent cells for temporal learning, 
  while placing relatively less emphasis on the complex spatial contents. In contrast, Fig.~\ref{fig:cate-b} 
  depicts models based on purely 2D convolutional neural networks (CNNs), where multiple frames are stacked over 
  the channel dimension~\cite{gaoSimVPSimplerBetter2022, tanTemporalAttentionUnit2023}. 
  This allows the usage of well-developed techniques for images and provides finer spatial granularity, 
  but lacks insights into the distinctiveness of temporal information. Different from them, 
  Video Vision Transformer~(ViViT)~\cite{arnabViViTVideoVision2021} or 3D-CNN-based models, 
  as shown in Fig.~\ref{fig:cate-c}, basically operate on 3D tokens (or patches) in the space and time 
  dimensions simultaneously~\cite{aignerFutureGANAnticipatingFuture2018, gaoEarthformerExploringSpacetime2022, ningMIMOAllYou2023a}. They exhibit promising performance due to the capability of perceiving spatiotemporal correlations at the same time. However, the spatial and temporal contents within those tokens are severely entangled since they are treated equivalently, which might not align with the intrinsic nature of physical laws in the real world.

  As a result, the resultant models in previous literature have their own focuses, and thus may excel at some video prediction scenarios but perform poorly on other ones.  Towards a generic video prediction solution, we argue that all the aforementioned aspects, \textit{i.e.}, spatial modeling, temporal modeling, and simultaneous spatiotemporal modeling, are essential. Although one can extend existing network architectures to model all these features, such as combining ViViT blocks with RNNs and CNNs, the developed models may have high computational costs and become complicated. 

  To explicitly model spatial, temporal, and spatiotemporal features in an efficient manner, we  propose a \textbf{SI}mple yet effective  \textbf{A}lternating \textbf{M}ixer (\methodname) for generic video prediction. 
The novelty of \methodname lies in the design of dimension alternating mixing (\blockname) blocks, which is composed of three types of Mixers: the spatial Mixer, the spatiotemporal Mixer, and the temporal Mixer, corresponding to the information extraction of spatial, spatiotemporal, and temporal features. 
Specifically, the spatial Mixer adopts depthwise convolutions and acts on the space and channel dimensions, responsible for updating the spatial appearances of individual frames; the spatiotemporal Mixer processes video tensors by 3D convolutions with the Inception style~\cite{yuInceptionNeXtWhenInception2023} to consider motion dynamics within local receptive fields; the temporal Mixer
utilizes multilayer perceptrons (MLPs) to boost the temporal evolution of video clips. We stack several \blockname blocks sequentially in the latent space formed by an encoder-decoder framework, alternating feature mixing of different levels. Despite its simplicity, the proposed \methodname benefits from a comprehensive understanding of videos and thus yields promising results in various scenarios.

\paragraph{Contributions.}The main contributions of this paper are summarized as follows. 1) We propose a simple yet effective video prediction model, \methodname, which can enjoy the best of three worlds---RNNs, 2D CNNs, and ViViTs/3D CNNs.
2) The proposed \methodname block can efficiently and effectively model spatial, temporal, and spatiotemporal features through alternating the dimensions of feature maps. 
3) Extensive experimental results  demonstrate the superior performance of the proposed \methodname over existing state-of-the-art methods on four benchmark video datasets.

\section{Related work}
\noindent \textbf{RNN-based models.}\quad Since video prediction can be interpreted as a sequence-to-sequence problem with frames as the basic units, the application of novel recurrent structures is natural. The pioneering ConvLSTM~\cite{shiConvolutionalLSTMNetwork2015b} extends the fully connected RNNs with vanilla convolutions. PredRNN~\cite{wangPredRNNRecurrentNeural2023} proposes the ST-LSTM unit and the spatiotemporal memory flow to keep state transitions consistent at different levels. PhyDNet~\cite{guenDisentanglingPhysicalDynamics2020} introduces the PhyCell to impose physical constraints on its memory flow, while MotionRNN~\cite{wuMotionRNNFlexibleModel2021} learns the transient variations and the trending momentum to represent motions by utilizing MotionGRU cells. The recent SwinLSTM~\cite{tangSwinLSTMImprovingSpatiotemporal2023} applies the multi-head self-attention mechanism to provide its recurrent backbone with more global information. Despite their in-depth explorations of temporal relationships, RNN-based models place less emphasis on the excavation of static spatial contents, which degrades the sharpness of the generated frames especially for long-term predictions.

\noindent \textbf{2D-CNN-based models}\quad This seems to be a straightforward design. It takes time as another form of channel and simply integrates them together so that techniques dedicated for images can be directly applied. Consequently, these models are able to comprehend spatial interactions more effectively. SimVP~\cite{gaoSimVPSimplerBetter2022} presents a purely 2D convolutional baseline and shows competitive performance as more complicated networks. It is further improved with the Temporal Attention Unit (TAU)~\cite{tanTemporalAttentionUnit2023} to focus on both intra- and inter-frame correlations, which leads to state-of-the-art results. However, visible artifacts are sometimes produced due to the lack of explicit temporal constrains.

\noindent \textbf{ViViT- and 3D-CNN-based models.}\quad Recent advances in ViViTs and 3D CNNs for video processing show their potential in the realm of video prediction: FutureGAN~\cite{aignerFutureGANAnticipatingFuture2018} builds a 3D convolutional network to perceive spatial and temporal contents at the same time. EarthFormer~\cite{gaoEarthformerExploringSpacetime2022} proposes a hierarchical Transformer based on Cuboid Attention, where the input tensor is decomposed into cuboids with a certain span in the height, width, and time. MIMO-VP~\cite{ningMIMOAllYou2023a} extends the ViViT with 3D local spatiotemporal blocks to grasp the short-term variation while maintaining the order information. However, we observe that video prediction necessitates an extension in time rather than space, which might conflict with these models since they inherently treat time and space equally. Besides, the high demand for computational resources appears to be another challenge for them.

\begin{figure*}[htb]
    \centering
  \includegraphics[width=.99\linewidth]{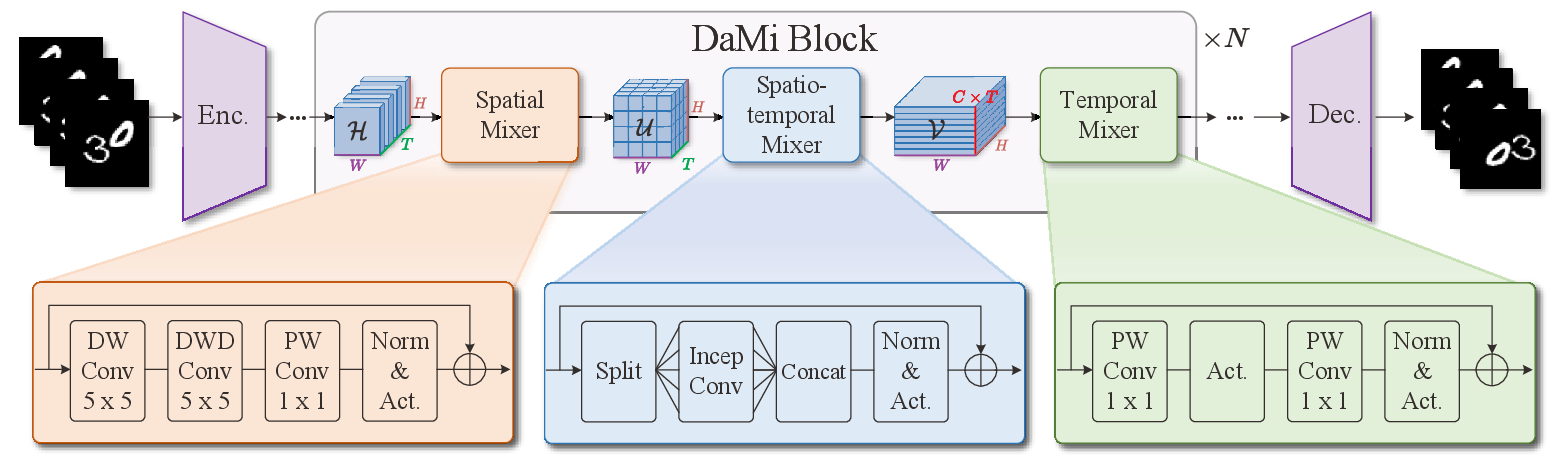}
  %
  \caption{The overall architecture of the proposed \methodname and the detailed structure of the \blockname block.}
  \label{fig:model}
  \end{figure*}

\section{Methodology}
In this section, we introduce \methodname, which alternates feature mixing in the latent space. We employ $N$ \blockname blocks between an encoder and a decoder, and each of them is composed of three types of Mixers to explicitly perceive spatial, temporal, and spatiotemporal correlations.
\subsection{Overall Architecture}
Fig.~\ref{fig:model} presents the overview of our \methodname. Inspired by SimVP~\cite{gaoSimVPSimplerBetter2022} and Latent Diffusion Models~\cite{rombachHighResolutionImageSynthesis2022a}, we attempt to learn the mapping from the latent representations of past frames to the future ones, which enables \methodname to focus on semantic dynamics and also reduces computational demands. 
  
Specifically, suppose that we are given an input sequence of $T$ frames, $\mathcal{X}_\mathrm{in} = \left( \boldsymbol{X}_1, \boldsymbol{X}_2, \ldots, \boldsymbol{X}_T \right)$, where $\boldsymbol{X}_t \in \mathbb{R}^{C \times H \times W}$ denotes the $t$-th frame with channel $C$, height $H$, and width $W$. 
The encoder encodes $\boldsymbol{X}_t$ into its latent representation $\boldsymbol{Z}_t = \textrm{Enc.}\left(\boldsymbol{X}_t\right) \in \mathbb{R}^{C^\prime \times H^\prime \times W^\prime}$, yielding $\mathcal{Z}_\mathrm{in} = \left( \boldsymbol{Z}_1, \boldsymbol{Z}_2, \ldots, \boldsymbol{Z}_T \right)$. 
This embedded video tensor $\mathcal{Z}_\mathrm{in} \in \mathbb{R}^{T \times C^\prime \times H^\prime \times W^\prime}$ is fed into the translation module, passing through $N$ \blockname blocks. As a result, it is converted into $\mathcal{Z}_\mathrm{out} = \textrm{Trans.}\left(\mathcal{Z}_\mathrm{in}\right) \in \mathbb{R}^{T^\prime \times C^\prime \times H^\prime \times W^\prime}$, representing the embeddings of the $T^\prime$ subsequent frames.
Finally, the decoder upsamples and reconstructs all predicted pictures $\mathcal{X}_\mathrm{out} = \left( \boldsymbol{X}_{T+1}, \boldsymbol{X}_{T+2}, \ldots, \boldsymbol{X}_{T+T^\prime} \right)$ from $\mathcal{Z}_\mathrm{out}$ in parallel. It should be noted that the encoder and the decoder consist of 2D convolutional layers and process each frame individually, while the translation module leverages multiple \blockname blocks to alternatively learn spatiotemporal dynamics.

\subsection{\blockname Block}
\label{subsec:mixers}
The \blockname block, which originates from MLP-Mixer~\cite{tolstikhinMLPMixerAllMLPArchitecture2021a} that contains channel mixing and token mixing operations, is the basic building unit of the translation module. The designs of its three components are tailor-made for a comprehensive modeling of video dynamics: the spatial Mixer, the spatiotemporal Mixer, and the temporal Mixer, as shown in Fig.~\ref{fig:model}. They are alternated to realize full awareness of videos from different perspectives, and thus give more confident predictions.

\noindent \textbf{Spatial Mixer.}\quad The spatial Mixer is responsible for learning spatial features within each frame by mixing them up in the space and channel dimensions.
Since depthwise convolutions give promising results in recent works~\cite{tanTemporalAttentionUnit2023}, we employ a depthwise convolution ($\texttt{DWConv}$), a depthwise dilated convolution ($\texttt{DWDConv}$), and a pointwise convolution ($\texttt{PWConv}$) to maintain a large receptive field while reducing computational costs. Specifically, the spatial Mixer separates the input tensor $\mathcal{H} \in \mathbb{R}^{T \times C^{\prime} \times H^{\prime} \times W^{\prime}}$ into $T$ individual frames and performs frame-wise operations as follows:
  \begin{align}
      \boldsymbol{U}_{t} &= \sigma\left(\texttt{Norm}\left(\boldsymbol{U}_{t}^{\prime}\right)\right) + \boldsymbol{H}_{t}\\
      &\text{where}\ \boldsymbol{U}_{t}^{\prime} = \texttt{PWConv}\left(\texttt{DWDConv}\left(\texttt{DWConv}\left(\boldsymbol{H}_{t}\right)\right)\right), \notag
  \end{align}
where $\boldsymbol{H}_t \in \mathbb{R}^{C^{\prime} \times H^{\prime} \times W^{\prime}}$ represents the $t$-th frame of $\mathcal{H}$; $\sigma(\cdot)$ is a non-linear activation function, which is ReLU in this paper. After being processed independently, all frames are concatenated to form the output tensor $\mathcal{U} \in \mathbb{R}^{T \times C^{\prime} \times H^{\prime} \times W^{\prime}}$.

  \noindent \textbf{Spatiotemporal Mixer.}\quad Inspired by InceptionNeXt~\cite{yuInceptionNeXtWhenInception2023}, we develop 3D depthwise convolutions with the Inception style, denoted as $\texttt{IncepConv}$, to build the spatiotemporal Mixer which aims to learn spatiotemporal correlations within each channel. The input tensor $\mathcal{U}$ is first split into five groups along the channel dimension, and then the $\texttt{IncepConv}$ layer applies convolutions with various kernels to them except for an identity mapping. Finally, all these groups are concatenated back together to form the output $\mathcal{V}$. This process can be formulated as follows:
  \begin{align}
          \mathcal{V} & = \sigma\left(\texttt{Norm}\left(\mathcal{V}^{\prime}\right)\right) + \mathcal{U}\\
          &\text{where}\ \mathcal{V}^{\prime} = \texttt{Concat}\left(\texttt{IncepConv}\left(\texttt{Split}\left(\mathcal{U}\right)\right)\right). \notag
  \end{align}

  \noindent \textbf{Temporal Mixer.} The temporal Mixer is to mix features across the time and channel dimensions, which concentrates on the feature evolution happening at a specific position throughout the entire span of time. Following MetaFormer~\cite{yuMetaFormerActuallyWhat2022}, we reshape the input  $\mathcal{V} \in \mathbb{R}^{T \times C^{\prime} \times H^{\prime} \times W^{\prime}}$ into $\tilde{\boldsymbol{V}}  \in \mathbb{R}^{\left(TC^{\prime}\right) \times H^{\prime} \times W^{\prime}}$ by stacking all frames along the channel dimension, and pass it through a two-layered MLP which maps $\mathbb{R}^{TC^\prime} \mapsto \mathbb{R}^{T C^\prime}$:
  \begin{align}
      \tilde{\boldsymbol{H}} =  \sigma\left(\texttt{Norm}\left(\mathbf{W}_2\,\sigma\left(\mathbf{W}_1\, \tilde{\boldsymbol{V}}\right)\right)\right)+ \tilde{\boldsymbol{V}},
      \label{eq:temporal}
  \end{align}
where $\mathbf{W}_1 \in \mathbb{R}^{r\left(TC^\prime\right)\times \left(TC^\prime\right)}$ and $\mathbf{W}_2 \in \mathbb{R}^{\left(TC^\prime\right) \times r\left(TC^\prime\right)}$ are learnable parameters with $r$ being the expansion ratio.
Note that the operations mentioned here are performed on the entire stacked ``image'', which are different from those performed on a single frame in the spatial Mixer.

  \section{Experiments}
  \label{sec:experiments}
  
  \subsection{Experimental Setup}
  \paragraph{Datasets.} 
  We evaluate our method on four benchmark datasets following \cite{tan2023openstl}, including
  1)~the widely used Moving MNIST (M-MNIST), 2)~TaxiBJ and 3)~WeatherBench involving complex physical fields, and 4)~the relatively high-resolution Human3.6M captured by real-world cameras. 
  \textbf{M-MNIST} contains two digits which are initialized at random locations, move with random velocities and bounce off the boundaries at certain angles.
  \textbf{TaxiBJ} records trajectories of traffic flow in Beijing, showing the inflow and outflow of each region at its corresponding pixel with two channels.
  \textbf{WeatherBench} contains physical quantities like temperature, humidity, and wind. According to~\cite{tan2023openstl}, we regrid the raw data with a spatial resolution of ${5.625}^\circ$ and choose the temperature field as the predictand.
  \textbf{Human3.6M} is a high-resolution RGB video dataset containing intricate human poses from the real world of 17 different scenarios. 
  Since videos of arbitrary length can be predicted in an autoregressive manner~\cite{voletiMCVDMaskedConditional2022}, we select the same length of input and output videos as previous works~\cite{gaoSimVPSimplerBetter2022,tanTemporalAttentionUnit2023} for fair comparison.

  \paragraph{Implementation details.} In this paper, all models are trained with $L_2$ loss. Pixel values of the input and target frames are normalized to $\left[0, 1\right]$. We choose $N=8$ \blockname blocks to form the translation module, and the expension ratio of the MLP in Eq.~\eqref{eq:temporal} is set to $r = 4$.  More detailed setups for each dataset are listed in Table~\ref{tab:setups}.

  \begin{table}[htb]
    \centering
    \caption{Experimental setups. $T_\mathrm{in/out}$: the length of the input/output video. \textbf{Frame shape}: the shape of a single input frame. \textbf{Latent shape}: the shape of an encoded frame. \textbf{Mixer dim}: the hidden dimension for each kind of Mixers.\loose}
    \resizebox{0.49\textwidth}{!}
      {\begin{tabular}{ccccc}
      \shline
      \rule{0pt}{8.5pt}
     & M-MNIST & TaxiBJ & WeatherBench & Human3.6M      \\      
      \midrule
      $T_\mathrm{in} \rightarrow T_\mathrm{out}$ & $10\rightarrow10$ & $4\rightarrow4$& $12\rightarrow12$ & $4\rightarrow4$  \\ [0.8ex]
      Frame shape & $\left(1, 64, 64\right)$& $\left(2, 32,32\right)$ & $\left(1, 32,64\right)$ & $\left(3, 256, 256\right)$  \\[0.8ex]
      Latent shape & $\left(64, 16, 16\right)$& $\left(64, 16,16\right)$ & $\left(32, 16,32\right)$ & $\left(128, 64, 64\right)$  \\[0.8ex]
       Mixer dim & $\left(256,256,640\right)$& $\left(64, 64, 256\right)$ & $\left(32, 32, 384\right)$ & $\left(512, 512, 512\right)$  \\ [0.5ex]

    \shline
       
      \end{tabular}}
      \vspace{-\baselineskip}
    \label{tab:setups}
  \end{table}

\begin{figure}[t]

  \centerline{\includegraphics[width=\linewidth]{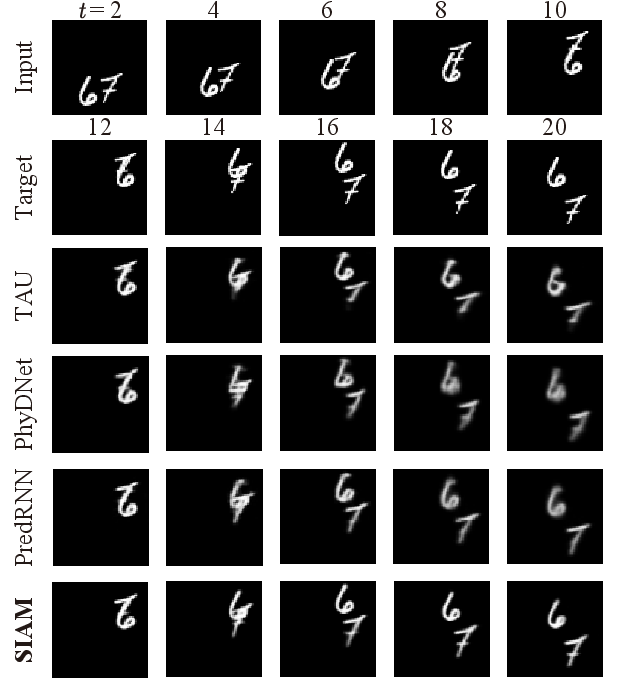}}

\caption{Predicted results on the M-MNIST dataset.}
\vspace{-5mm}

\label{fig:mmnist}
\end{figure}

\subsection{Qualitative Evaluation}

We first compare qualitative results of \methodname with other approaches on M-MNIST. As shown in Fig.~\ref{fig:mmnist}, the two digits meet each other at $t=10$ and are intertwined from $t=12$ to $t=14$, which poses a great challenge for long-term predictions. Most of the frames produced by the compared models are not very satisfactory: PhyDNet~\cite{guenDisentanglingPhysicalDynamics2020} and TAU~\cite{tanTemporalAttentionUnit2023} generate blurry images, failing to reconstruct the digit `6' or `7' which becomes more distorted and illegible as the time progresses; PredRNN~\cite{wangPredRNNRecurrentNeural2023} performs relatively well, but loses details like the horizontal line through the middle of the digit `7' at $t=18$ and $t=20$. Such setbacks highlight the limited perception of different perspectives inherent in these methods. In contrast, our model produces clearer images, accurately predicting the trajectories of the moving digits while maintaining their shapes after occlusions. This suggests that \methodname can effectively capture both the temporal dynamics and the static spatial content, which translates to its superior performance in complex scenarios. This robust prediction capability is largely attributed to its comprehensive and integrated understanding of video data by alternating Mixers inside the \blockname blocks.
\begin{figure}[t]

      \centerline{\includegraphics[width=\linewidth]{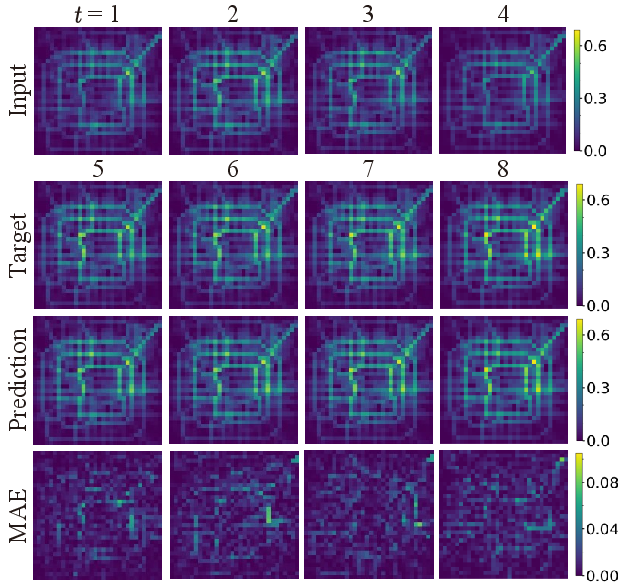}}

    \caption{Predicted results on the TaxiBJ dataset. MAE denotes the absolute error between the predicted results and the corresponding targets.}
    \vspace{-5mm}
    \label{fig:taxibj}
    \end{figure}
    
Fig.~\ref{fig:taxibj} visualizes an example of our method predicting the traffic flow on TaxiBJ. In this case, the intensity of the traffic flow increases at $t=5$, which may correspond to the unexpected congestion during early morning hours. Despite this sudden change of the traffic flow, the proposed \methodname can still grasp the overall trend and predict the future with high fidelity, demonstrating its flexibility granted by the three kinds of Mixers for alternative feature extractions. Moreover, these accurate predictions on the TaxiBJ dataset further suggest its potential applications in the real world, such as traffic estimation and management.

Please refer to Appendix~\ref{supp:qualitative} in the supplementary material for  more visualization results on these datasets.
    
\begin{table}[tb]
    \centering
    \caption{Comparisons with the state-of-the-arts. The optimal (suboptimal) results are marked by bold (underlined).\loose}
    \resizebox{0.49\textwidth}{!}
      {\begin{tabular}{crrrrrr}
      \shline
      \rule{0pt}{12.5pt}
      Dataset & Method & \makecell{\#Params.\\(M)} & \makecell{FLOPs\\(G)} & \makecell{MSE\\($\downarrow$)}   & \makecell{MAE\\($\downarrow$)}   & \makecell{SSIM\\($\uparrow$)} \\
      \midrule
      \multirow{8}[2]{*}{\rotatebox{90}{M-MNIST}} 
            & ConvLSTM~\cite{shiConvolutionalLSTMNetwork2015b} & 15.1    & 56.8  & 34.4  & 98.0  & 0.898 \\ %
            & MAU~\cite{changMAUMotionAwareUnit2021} & 4.5  & 17.8   & 24.0  & 72.0  & 0.937 \\ %
            & PredRNN~\cite{wangPredRNNRecurrentNeural2023} & 23.9  & 116.6   & 23.9  & 71.7  & 0.936 \\ %
            & PhyDNet~\cite{guenDisentanglingPhysicalDynamics2020} & 3.1   & 15.3  & 23.3  & 68.0  & 0.940 \\ %
            & SimVP~\cite{gaoSimVPSimplerBetter2022} & 58.0    & 19.4  & 26.5  & 76.9  & 0.930 \\ %
            & TAU~\cite{tanTemporalAttentionUnit2023}  & 44.7 & 15.6 & \underline{20.0}  & \underline{62.0}   & 0.948 \\ %
            & SwinLSTM~\cite{tangSwinLSTMImprovingSpatiotemporal2023} & 20.2 & 69.6 & 20.9 & 64.0 & \underline{0.954} \\ %
            & \methodname~(\textbf{ours}) & 34.6 & 16.4 & \textbf{17.3}  & \textbf{55.4} & \textbf{0.962} \\
      \midrule
      \multirow{8}[2]{*}{\rotatebox{90}{TaxiBJ}} & ConvLSTM~\cite{shiConvolutionalLSTMNetwork2015b} & 15.0 & 20.7 & 0.336 & 15.4  & 0.977 \\ %
            & MAU~\cite{changMAUMotionAwareUnit2021} & 4.4 & 6.6 & 0.344 & 15.5 & 0.976 \\ %
            & PredRNN~\cite{wangPredRNNRecurrentNeural2023}& 23.7 & 42.6 & 0.375 & \underline{15.1}  & 0.977 \\ %
            & PhyDNet~\cite{guenDisentanglingPhysicalDynamics2020}& 3.1  & 5.6   & 0.371 & 15.7  & 0.975 \\ %
            & SimVP~\cite{gaoSimVPSimplerBetter2022} & 13.8 & 3.6  & \underline{0.331}  & 15.2  & 0.977 \\ %
            & TAU~\cite{tanTemporalAttentionUnit2023}  & 9.6  & 2.5  & 0.333 & \underline{15.1}  & \underline{0.978} \\ %
            & SwinLSTM~\cite{tangSwinLSTMImprovingSpatiotemporal2023}  & 11.1 & 4.7 & 0.354 & 15.8 & 0.977 \\ %
            & \methodname~(\textbf{ours}) & 4.0 & 1.2 & \textbf{0.301} & \textbf{14.7}  & \textbf{0.981} \\ %
    \midrule
      \multirow{8}[2]{*}{\rotatebox{90}{WeatherBench}} & ConvLSTM~\cite{shiConvolutionalLSTMNetwork2015b} & 15.0 & 135.7   & 82.7  & 265.3 & 0.982 \\ 
            & MAU~\cite{changMAUMotionAwareUnit2021} & 5.5  & 39.6   & 69.1  & 237.0  & 0.986 \\ 
            & PredRNN~\cite{wangPredRNNRecurrentNeural2023} & 23.6 & 277.5   & 72.4  & 242.0 & 0.984 \\
            & MIM~\cite{wangMemoryMemoryPredictive2019c}  & 37.8 & 108.7   & 86.0  & 269.4 & 0.983 \\ 
            & SimVP~\cite{gaoSimVPSimplerBetter2022} & 14.7 & 8.0  & 68.5  & 237.0 &  0.985\\ %
            & TAU~\cite{tanTemporalAttentionUnit2023}   & 12.2 & 6.7   & \underline{66.9}  & \underline{236.7} & 0.985 \\ %
            & SwinLSTM~\cite{tangSwinLSTMImprovingSpatiotemporal2023}  &  11.1 & 122.0 & 70.2 & 241.1 & \underline{0.986} \\
            & \methodname~(\textbf{ours}) & 9.6 & 6.0 & \textbf{64.2}  & \textbf{225.3} & \textbf{0.990} \\ %
      \midrule
      \multirow{8}[2]{*}{\rotatebox{90}{Human3.6M}} & ConvLSTM~\cite{shiConvolutionalLSTMNetwork2015b} & 15.5 & 346.5 & 125.5 & 1561  & 0.964 \\ %
            & MAU~\cite{changMAUMotionAwareUnit2021} & 20.2 & 104.9 & 130.6 & 1622 & 0.962\\
            & PredRNN~\cite{wangPredRNNRecurrentNeural2023}& 24.6 & 707.9 & 115.7 & 1488  & 0.966 \\ %
            & MIM~\cite{wangMemoryMemoryPredictive2019c} & 47.6 & 1050.6 & 117.4 & 1468 & \textbf{0.969} \\ %
            & SimVP~\cite{gaoSimVPSimplerBetter2022} & 41.2 & 197.2  & 116.2  & 1512 & 0.966 \\ %
            & TAU~\cite{tanTemporalAttentionUnit2023}  &  37.6 & 181.8  & \underline{114.1} & 1587  & \underline{0.968} \\ %
            & SwinLSTM~\cite{tangSwinLSTMImprovingSpatiotemporal2023}  &  11.1 & 297.6 & 117.0 &\underline{1467} &0.967\\ %
            & \methodname~(\textbf{ours}) & 24.0 & 180.5 & \textbf{105.5} & \textbf{1435}  & \textbf{0.969} \\ %

        \shline
      \end{tabular}}
    \label{tab:quantitative}
  \end{table}

\subsection{Quantitative Evaluation}
In Table~\ref{tab:quantitative}, we evaluate our method against several strong baselines using the frame-wise Mean Squared Error (MSE), Mean Absolute Error (MAE), and Structural Similarity Index Measure (SSIM). Parameters and FLOPs are also counted as estimations of model complexity and efficiency. We see that \methodname consistently outperforms recent state-of-the-art methods on all benchmark datasets, giving more reliable predictions with less deviation from the ground truth. In particular, \methodname reduces the MSE of currently best TAU~\cite{tanTemporalAttentionUnit2023} on M-MNIST from $20.0$ to $17.3$ ($13.5\%\downarrow$), and also brings an improvement of $10.5\%$ for the MAE metric. For the complicated Human3.6M recorded by cameras, it exceeds its competitors over $7.5\%$ in terms of the MSE. Meanwhile, it is more parameter-efficient than SimVP~\cite{gaoSimVPSimplerBetter2022} and TAU~\cite{tanTemporalAttentionUnit2023}, and consumes less computational resources than RNN-based approaches like MAU~\cite{changMAUMotionAwareUnit2021} and SwinLSTM~\cite{tangSwinLSTMImprovingSpatiotemporal2023}. 
This is primarily ascribed to the strategic design of the latent space and the use of alternative Mixers for multidimensional modeling.

\begin{table}[htb]
    \centering
    \caption{Ablation studies of different variants of \blockname on the M-MNIST dataset.}

    \resizebox{0.49\textwidth}{!}{
    \begin{tabular}{cccccrrrr}
        \shline
        \rule{0pt}{12.5pt}
     & \makecell{Exp.\\index} & \makecell{Spatial \\Mixer}  & \makecell{Spatiotemporal \\Mixer}  & \makecell{Temporal \\Mixer}  & \makecell{MSE\\($\downarrow$)}& \makecell{MAE. \\ ($\downarrow$)} & \makecell{SSIM \\ ($\uparrow$)}  \\
     \midrule
    \multirow{3}{*}{\makecell{Single\\Mixer}} & (a) & \Checkmark & \XSolidBrush & \XSolidBrush &  148.9 & 330.5 & 0.322 \\
    & (b) &  \XSolidBrush &  \Checkmark & \XSolidBrush & 32.3 & 89.0 & 0.914 \\
    & (c) &  \XSolidBrush & \XSolidBrush & \Checkmark   &   55.2  &   140.0   & 0.817\\
    \midrule
    \multirow{3}{*}{\makecell{Dual\\Mixers}} & (d) & \Checkmark   & \Checkmark   & \XSolidBrush &  22.5   &  67.7     & 0.942 \\
    & (e) & \Checkmark   & \XSolidBrush & \Checkmark   &   18.6  &  59.6    & 0.951\\ 
    & (f) & \XSolidBrush & \Checkmark   & \Checkmark   &   19.3  &  60.4    & 0.950 \\
    \midrule
    \blockname & (g) & \Checkmark   & \Checkmark   & \Checkmark   &  \textbf{17.3}   & \textbf{55.4}    &  \textbf{0.962}\\
    \shline
    \end{tabular}}
    \label{tab:ablation}
    \end{table}
    
\subsection{Ablation Study}
In order to validate our motivation and the design of \blockname, we carry out ablation studies on M-MNIST to see how \blockname blocks work in the latent space. Comparisons of experimental results in various settings are summarized in Table~\ref{tab:ablation}.

\noindent \textbf{Single Mixer.}\quad 
We assess the predictive ability of each kind of Mixers in Exp.~(a)-(c). Obviously, none of them gives a satisfactory result when applied separately. However, we observe that the spatiotemporal Mixer (Exp.~(b)) performs relatively better compared to the others since it considers both spatial and temporal aspects. It can be inferred that perceiving a video from a fixed viewpoint is far from sufficient due to the absence of multidimensional information, \textit{e.g.}, temporal consistency for Exp.~(a) and spatial contents for Exp.~(c). 

\noindent \textbf{Dual Mixers.}\quad
Exp.~(d)-(f) harness various combinations of two Mixers as the building blocks of the translation module. They exhibit competitive performance even with the previous state-of-the-arts, and also surpass the single-Mixer ones by large margins. This indicates that these Mixers work in a complementary way to ameliorate the quality of generated frames. Moreover, Exp.~(e), which models temporal and spatial features separately before fusing them, yields better results than simultaneously modeling spatiotemporal features via the spatiotemporal Mixer (Exp.~(b), (d), and (f)). This observation supports our motivation that all aspects of video representations contribute to accurate predictions.

\noindent \textbf{\blockname with triplets of Mixers.}\quad
Although the dual-Mixer variants have offered efficient and effective baselines, the performance can be further refined by integrating the spatial Mixer, the spatiotemporal Mixer,  and the temporal Mixer together as shown in Exp.~(g). Compared to Exp.~(f), the MSE is improved significantly ($10.4\%\downarrow$) while the complexity growth ($2.7\%\uparrow$) is still controllable. This fairly good promotion is attributed to the explicit feature modeling facilitated by \blockname blocks, and the decomposition of video dimensions via those Mixers also guarantees computational efficiency.

Please refer to Appendix~\ref{supp:abla} in the supplementary material for more ablation studies on serial orders and parallel connections of these three Mixer modules.

\section{Conclusion}
In this paper, we introduce \methodname, a simple yet effective framework leveraging Mixers for video prediction. Our method learns the evolution from past frames to future ones in the latent space of an autoencoder. Different types of Mixers are alternated for explicitly modeling the features in all of the spatial, temporal, and spatiotemporal aspects.
Extensive experimental results demonstrate the superior performance and also the efficiency of \methodname over existing methods. In future work,  it would be intriguing to explore more attentive Mixers, thereby extending the utility of  \methodname across diverse tasks.

\bibliographystyle{IEEEtran}
\bibliography{ref}

\clearpage
\setcounter{page}{1}
\appendices

   \newpage
       \twocolumn[
        \centering
        \Large
        \textbf{\methodname: A Simple Alternating Mixer for Video Prediction}\\
        \vspace{0.5em}Supplementary Material \\
        \vspace{1.0em}
       ] 

\counterwithin{table}{section} 
\counterwithin{figure}{section} 
\renewcommand{\thetable}{\Alph{section}\arabic{table}} 
\renewcommand{\thefigure}{\Alph{section}\arabic{figure}} 

\section{Ablation study on Mixer connections}
\label{supp:abla}
The proposed \blockname block is composed of the spatial Mixer, the spatiotemporal Mixer, and the temporal Mixer to comprehensively learn video features of different aspects. Actually, how the three types of Mixers form a \blockname block can be various since we have the flexibility to sequentially alternate them in different orders, or parallelize their connections. Therefore, we conduct experiments on the M-MNIST dataset to investigate the best structure of the \blockname block.

Table.~\ref{tab:abla_connect} shows the quantitative results for this ablation. While each implementation of the \blockname block has its own strengths and weaknesses with respect to different metrics, the overall performance differences are minimal. This suggests that the way Mixers are connected is less critical to prediction accuracy. Since the parallel one consumes more GPU memories, we select the best serial setup, \textit{i.e.}, alternating with the order of the spatial Mixer, the spatiotemporal Mixer, and the temporal Mixer, as described in the main paper.

\begin{table}[htb]
\centering
\caption{Ablation study of Mixer connections on the M-MNIST dataset.}

\resizebox{0.49\textwidth}{!}{
\begin{tabular}{ccccrrrr}
    \shline
    \rule{0pt}{12.5pt}
 &  \makecell{Spatial \\Mixer}  & \makecell{Spatiotemporal \\Mixer}  & \makecell{Temporal \\Mixer}  & \makecell{MSE\\($\downarrow$)}& \makecell{MAE. \\ ($\downarrow$)} & \makecell{SSIM \\ ($\uparrow$)}  \\
 \midrule
\multirow{6}{*}{Serial} 
&  1 & 2 & 3 & \textbf{17.3}   & \textbf{55.4}    &  \textbf{0.962}  \\
& 1 & 3 & 2 & 17.4 & 55.8& 0.961\\
 &  2 &  1 & 3 & 17.5 & 56.4 & 0.960 \\
 & 2 & 3 & 1 & 17.5& 55.7& 0.961\\
 & 3 & 1 & 2 & 17.6&56.8 &0.960 \\
 & 3 & 2 & 1 & 17.7 & 56.3 & 0.960\\
\midrule

Parallel & -   & -   & -   & \textbf{17.3} & 55.5  & \textbf{0.962}\\
\shline
\end{tabular}}
\label{tab:abla_connect}
\end{table}

\section{Additional qualitative results}
\label{supp:qualitative}
\begin{figure*}[htb]

  \centerline{\includegraphics[width=0.99\linewidth]{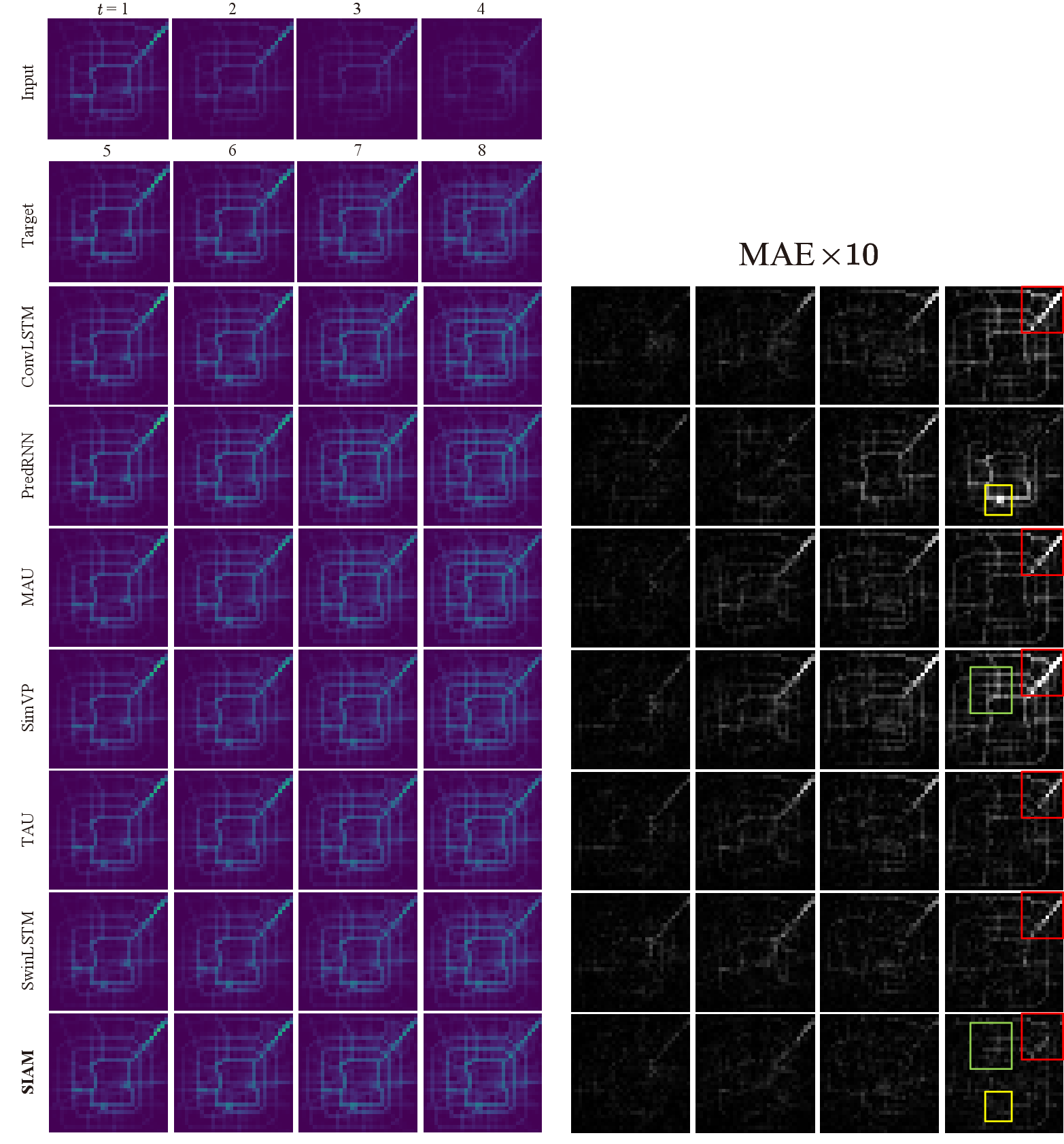}}

\caption{Comparisons with the state-of-the-arts on the TaxiBJ dataset. \textit{Left}: the predicted traffic flows. \textit{Right}: the scaled MAE between the predictions and the targets; darker areas represent regions with smaller errors.}
\label{supp:taxibj}
\end{figure*}

\begin{figure*}[htb]

  \centerline{\includegraphics[width=0.99\linewidth]{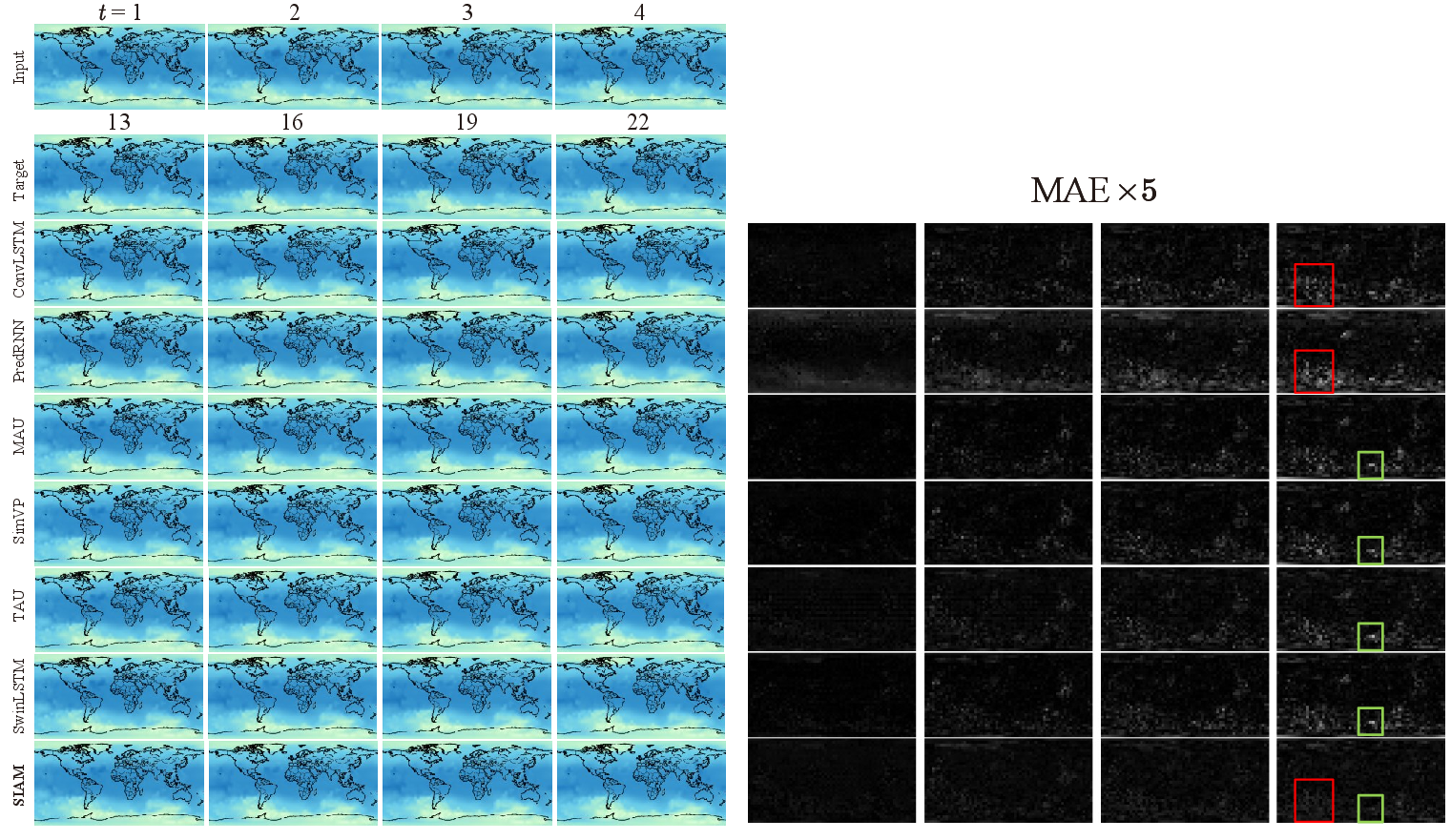}}

\caption{Comparisons with the state-of-the-arts on the WeatherBench dataset. \textit{Left}: the predicted temperature fields. \textit{Right}: the scaled MAE between the predictions and the targets; darker areas represent regions with smaller errors.}
\label{supp:weather}
\end{figure*}

\begin{figure*}[!h]

  \centerline{\includegraphics[width=0.99\linewidth]{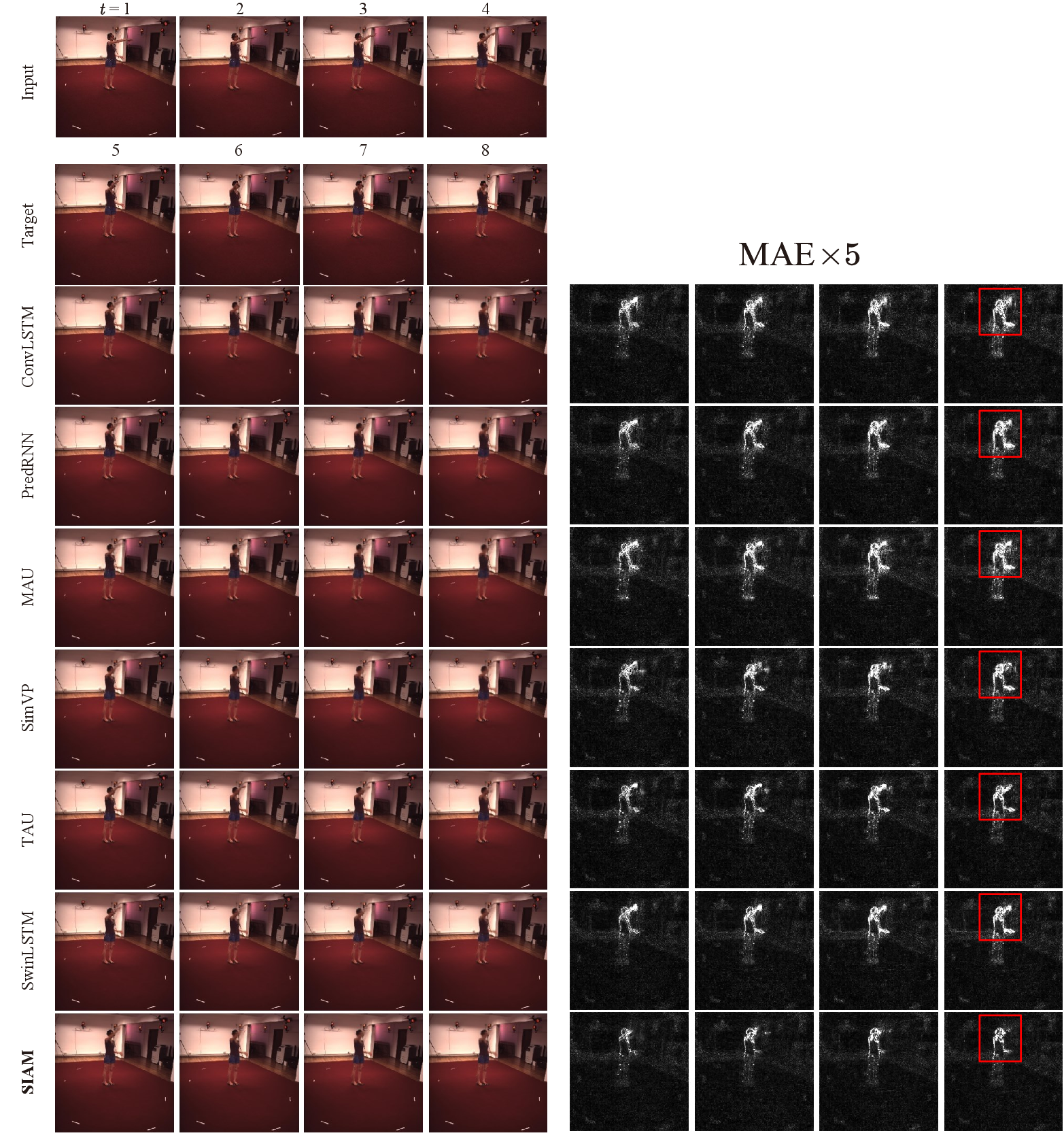}}

\caption{Comparisons with the state-of-the-arts on the Human3.6M dataset. \textit{Left}: the predicted human poses. \textit{Right}: the scaled MAE between the predictions and the targets; darker areas represent regions with smaller errors.}
\label{supp:human}
\end{figure*}

We provide additional results on qualitative comparisons with the state-of-the-arts on the M-MNIST, TaxiBJ, WeatherBench, and Human3.6M datasets. The scaled absolute errors are shown in the right parts of Fig.~\ref{supp:taxibj}, Fig.~\ref{supp:weather}, and Fig.~\ref{supp:human} to more clearly visualize the differences between  the predictions and their corresponding ground truths. 

\begin{figure}[t]

  \centerline{\includegraphics[width=\linewidth]{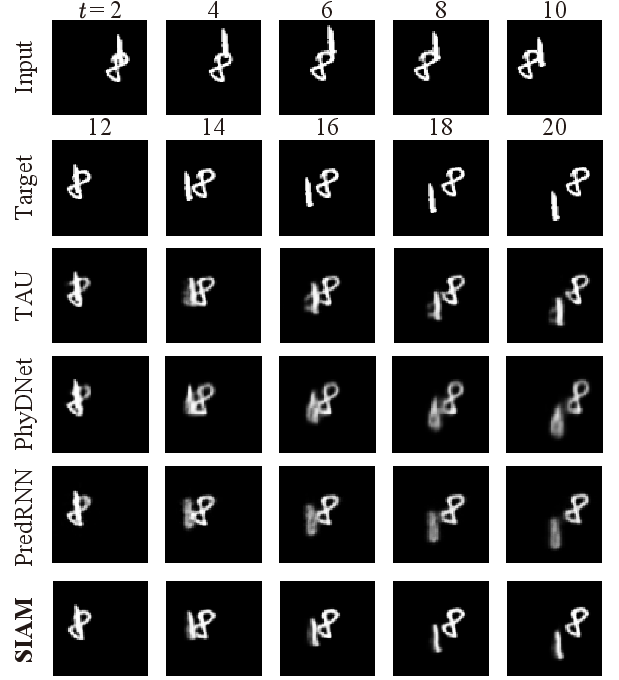}}
%
\caption{Predicted results on the M-MNIST dataset.}
\label{supp:mmnist}
\end{figure}

\noindent \textbf{M-MNIST.} \quad Fig.~\ref{supp:mmnist} presents visualizations on the M-MNIST dataset. The occlusion at $t=12$ makes it difficult to keep the sharpness of the predicted digits. We could see that \methodname keeps the shape of the digit `1' unchanged even until $t=20$, which further indicates the superior performance of the proposed \methodname.

\noindent \textbf{TaxiBJ.} \quad Fig.~\ref{supp:taxibj} presents the experimental results on a hard case of TaxiBJ, where the traffic flow first decreases and then increases. It is obviously that \methodname provides more accurate predictions than its competitors in the areas inside the red, green, and yellow boxes.

\noindent \textbf{WeatherBench.} \quad In Fig~\ref{supp:weather}, we present the global temperature forecasting results on the WeatherBench dataset. It can be seen that the frames generated by \methodname are quite similar to the ground-truth ones, and the MAE maps are also more sparse when compared to others, which suggests its potential applications in the real-world weather forecasting.

\noindent \textbf{Human3.6M.} \quad In Fig.~\ref{supp:human}, we present a clip of video from Human3.6M, where a young woman is lifting her right hand. In this case, it is challenging to predict the motion of the foreground object, \textit{i.e.}, the moving hand, while ensuring the static background objects remain unchanged. We could see that \methodname provides more accurate predictions of the moving hand with less deviation than others inside the red boxes.

\end{document}